\documentclass[a4paper]{article}

\usepackage{INTERSPEECH2021}
\usepackage{comment}
\usepackage{float}
\usepackage[utf8]{inputenc}
\usepackage{hyperref}
\usepackage{graphicx}
\usepackage{placeins}
\usepackage{multirow}
\usepackage{cleveref}
\usepackage{xcolor}

\makeatletter
\renewcommand{\section}{\@startsection
  {section}%
  {1}%
  {}%
  {-0.2\baselineskip}%
  {0.2\baselineskip}%
  {}}%

\renewcommand{\subsection}{\@startsection
  {subsection}%
  {2}%
  {}%
  {-0.1\baselineskip}%
  {0.1\baselineskip}%
  {}}%

\renewcommand{\subsubsection}{\@startsection
  {subsubsection}%
  {3}%
  {}%
  {-0.1\baselineskip}%
  {0.1\baselineskip}%
  {}}%

\g@addto@macro\normalsize{%
  \setlength\abovedisplayskip{5pt plus 2pt minus 2pt}
  \setlength\belowdisplayskip{5pt plus 2pt minus 2pt}
  \setlength\abovedisplayshortskip{4pt plus 2pt minus 2pt}
  \setlength\belowdisplayshortskip{4pt plus 2pt minus 2pt}
}

\makeatother

\title{The Impact of ASR  on the Automatic Analysis of Linguistic Complexity and Sophistication in Spontaneous L2 Speech}
\name{Yu Qiao$^1$, Wei Zhou$^{1,2}$, Elma Kerz$^1$, Ralf Schl\"uter$^{1,2}$}
\address{ $^1$RWTH Aachen University\\
        $^2$AppTek GmbH}
\email{yu.qiao@rwth-aachen.de, zhou@cs.rwth-aachen.de, elma.kerz@ifaar.rwth-aachen.de,  schlueter@cs.rwth-aachen.de}

\begin{document}

\maketitle
\begin{abstract}
In recent years, automated approaches to assessing linguistic complexity in second language (L2) writing have made significant progress in gauging learner performance, predicting human ratings of the quality of learner productions, and benchmarking L2 development. In contrast, there is comparatively little work in the area of speaking, particularly with respect to fully automated approaches to assessing L2 spontaneous speech. While the importance of a well-performing ASR system is widely recognized, little research has been conducted to investigate the impact of its performance on subsequent automatic text analysis. In this paper, we focus on this issue and examine the impact of using a state-of-the-art ASR system for subsequent automatic analysis of linguistic complexity in spontaneously produced L2 speech. A set of 30 selected measures were considered, falling into four categories: syntactic, lexical, n-gram frequency, and information-theoretic measures. The agreement between the scores for these measures obtained on the basis of ASR-generated vs. manual transcriptions was determined through correlation analysis. A more differential effect of ASR performance on specific types of complexity measures when controlling for task type effects is also presented. 
\end{abstract}


\section{Introduction}
Measuring the product of second language (L2) (spoken or written) performance is a key aspect of research into language learning and has a relatively long tradition dating back to 1970. This line of research has been primarily geared towards assessing L2 performance in terms of reliable and valid performance indicators that enable comparability and replicability across different studies (see e.g. \cite{larsen1978esl}). The search for such indicators has led to the emergence of the Complexity, Accuracy and Fluency (CAF) triad, a conceptual framework that captures the multicomponential nature of L2 proficiency (Housen \& Kuiken 2009, \cite{housen2012dimensions}, \cite{housen2019multiple}). \textit{Complexity} refers to the range, comprehensiveness, richness, and sophistication of language performance. \textit{Accuracy} refers to target-like and/or error-free language use. \textit{Fluency} typically refers to the smooth, easy, and eloquent production of language with a relatively small number of pauses or reformulations. 

As L2 learners become more proficient -- i.e. as compared to learners at lower levels or to themselves at earlier developmental levels -- they are expected to produce more complex and varied sentence structures and advanced vocabulary as well as more accurate and more fluent language. The significance of CAF is also reflected in the descriptors of language proficiency levels defined in European and international competence frameworks, such as the Common European Framework of Reference for Languages (CEFR, \cite{council2018common}) and its six levels of competence (A1-C2). Since the focus here is on complexity -- which in itself is multidimensional -- we provide below a concise overview of related work on automated assessment of indicators pertaining to this construct. At the lexical level, several measures have yielded consistent results in capturing differences associated with L2 English proficiency, such as those pertaining to lexical sophistication, lexical variation/diversity and lexical density (see e.g. \cite{lu2012relationship}, \cite{kyle2018tool}). Lexical sophistication refers to the degree to which a learner's production contains unusual or advanced words \cite{laufer1995vocabulary} and is commonly gauged using features related to word frequency, n-gram frequency, academic language, and psycholinguistic word properties, such as word prevalence. Lexical diversity refers to the degree to which the vocabulary used in L2 production is varied \cite{malvern2004lexical} and is often captured through features related to the number of word types produced in a language sample (the type-token ratio and its variants). Lexical density refers the amount of information in a learner text and is captured by the number of lexical (as opposed to grammatical) words to the total number of words in a language sample \cite{ure1971lexical}. At the syntactic level, a number of measures have produced mixed results in terms of L2 ability and development (see, e.g., \cite{norris2009towards}, \cite{bi2020syntactic}, see also \cite{crossley2020linguistic} for a recent review). Syntactic complexity measures can be broadly classified into three types: length-based, clausal subordination, and phrasal complexity measures. Length-based measures (e.g., number of words per sentence/utterance) are commonly used in studies of early first language acquisition based on the assumption that longer utterances are inherently more complex \cite{brown1973development}. However, sentence length measures alone are insufficient to assess later stages of L2 learning, as longer sentences do not always indicate more complex syntactic structure. Measures relating to clausal subordination are thus more commonly used in L2 studies for operationalizing syntactic complexity. In this way, clause structure is prioritized over length, so that a short sentence with multiple clauses is considered more complex than a longer sentence with a single clause. Measures of phrasal complexity, in particular noun phrase complexity, are typically associated with proficiency in academic writing, whereas clausal subordination is associated with proficiency in the spoken register (e.g. \cite{biber2016predicting}). 

In the domain of L2 writing, automated approaches to assessing complexity and sophistication have been successfully used to distinguish non-native from native and expert writers, to differentiate levels of L2 proficiency, and to predict human ratings of the quality of learner productions (see e.g. \cite{stroebel2016:cocogen}, \cite{kyle2018tool}; see also \cite{michel2017complexity}, \cite{lu2020automated}, \cite{crossley2020linguistic} for overviews). Comparatively little work has been undertaken in the area of speaking, in particular concerning fully automated approaches to assessment of L2 spontaneous speech, which includes measures of complexity and sophistication (however, for exceptions, see e.g. \cite{wang2018towards}). Such a fully automated approach requires automatic speech recognition (ASR) as a first step, which is known to have a higher error rate for spontaneous, non-native speech compared to predictive speech (for details on factors that may affect ASR accuracy, see \cite{zechner2020automated}: p.67). More generally, ASR of L2 speech is challenging due to a number of reasons: A key challenge is the lack of publicly available databases and benchmark datasets for spontaneously produced non-native speech. This is especially true for L2 speech produced by adolescents and young adults (but see the recent Interspeech (2020 \& 2021) shared tasks on ASR for Non-Native Children’s Speech). Another challenge is the considerable inter-individual variability among non-native speakers, even among those who are at similar levels of proficiency due to a host of factors, including speaking rate, vocal effort and speaking style. The phenomenon of idiosyncratic differences in ASR system performance has been known for many decades (see \cite{benzeghiba2007automatic} for a comprehensive review of ASR and speech variability), but its effects are more pronounced when the overall WER is higher, as is the case with spontaneous non-native speech. A third challenge associated with ASR of L2 spontaneous speech is the frequent occurrence of a number of phenomena that impair ASR performance, including mispronounced words, ungrammatical utterances and disfluencies (such as false starts, partial words, and filled pauses).

To our knowledge, while there is general recognition of the importance of a well-performing ASR in the assessment of spontaneous L2 speech, there is virtually no research aimed at understanding what role ASR accuracy plays in subsequent analyses of speech performance (for exceptions see \cite{tao2016exploring},\cite{knill2018impact}, \cite{lu2019impact}). In this paper, we focus on this issue and examine the impact of using a state-of-the-art (SOTA) ASR system for subsequent automatic analysis of linguistic complexity in spontaneously produced L2 speech. 
More specifically, the uncertainty introduced into the evaluation process by using automatically instead of manually transcribed speech is investigated.

\FloatBarrier
\vspace{-1mm}
\section{Data}
\vspace{-1mm}
The data used here come from two data sets of non-native (L2) spontaneous speech collected as part of the third author's research aimed at advancing our current understanding of L2 learning and development. The target population are L2 speakers of English with German as their first language at upper intermediate to advanced levels of proficiency. 

\textbf{School Data}:  The first data set consists of 165 samples of spontaneous L2 speech produced by as many students attending German high schools in North Rhine-Westphalia (NRW). This amounts to 9.7 hours of audio data. All students attended grades 10, 11, and 12, which generally correspond to ages 16-18. Students in grades 10 through 12 prepare for the high school diploma (Abitur), which is equivalent to the British A-level or the American high school diploma after 12 years of formal education. The core curricula for L2 English instruction at Gymnasien in NRW are oriented toward the CEFR framework and its proficiency levels mentioned in Section 1. According to these curricula, by the end of 10th grade students should reach level B1 of the CEFR with portions of level B2, while by the end of 12th grade they should reach level B2 with parts of level C1 in the receptive areas. The LexTALE task is a useful instrument designed for intermediate to advanced L2 learners and often used as a proxy estimate of general English proficiency (for more details see Lemhöfer \& Broersma 2012). 
The students' performance on the LexTALE task provided additional support for these proficiency levels as represented by mean score and standard deviation (SD): 10th grade = 64.25\% (SD 9.53), 11th grade = 67.83\% (SD 11.04) and 12th grade = 71.92\% (SD 12.19), where B2 = upper intermediate level corresponds to LexTale scores between  60\%-80\%. All students were asked to deliver a short (3-5 minutes) presentation on a key topic from the core curricula mentioned above that relates to the political system in the UK or UK international relations.  

\textbf{University Corpus}: The second  data set consists of 299 samples of spontaneous L2 speech produced by 243 students at RWTH Aachen University. 56 of them produced two speeches. This amounts to 26.5 hours of audio data. To elicit spontaneous speech, the students first watched a popular TED talk\footnote{\url{https://bit.ly/3cZw7fN}}
. They were then asked to present their views on some of the issues raised in the corresponding TED Talk. To ensure sufficiently sized speech samples (approx. 5 minutes), students received guiding questions to keep up the flow of speaking. Since the prerequisite for entry into the German university is the Abitur, students are expected to be at least B2 level and in parts C1 (see descriptions of the core curricula above). This is also reflected in their average LexTALE score of 75.79\% (SD 15.46).

The audio recording of the two data sets were manually transcribed following the same transcription conventions. Details of the data are shown in Table \ref{tab:distributions}. The train, dev and test sets have no speaker overlap.

\begin{table}[t]
    \centering
        \caption{Details of the L2 speech data}
    \begin{tabular}{lc|crr}
    \hline
         \multicolumn{2}{c|}{Data sets} & Recordings & Words & Hours \\
         \hline
School & dev & 19 & 7503  & 1.0\\
       & test& 50 & 23154 & 3.0\\
       &train& 96 & 42217 & 5.7\\
\hline
University &dev   &25  &13663 & 1.8\\
           &test  &33  &21584 & 2.9\\
           &train&240  &165861 & 21.8\\
\hline
    \end{tabular}
    \vspace{-3mm}
        \label{tab:distributions}
\end{table}

\FloatBarrier

\section{Automatic Speech Recognition Setup}
We use the hybrid hidden Markov Model (HMM)-based ASR system from \cite{zhou2020rwth} as our baseline, which shows SOTA performance on the 2nd release of TED-LIUM task (TLv2) \cite{tedlium2}. 
The bidirectional long short-term memory \cite{hochreiter1997lstm} (BLSTM)-based acoustic model (AM) is further fine tuned on the training sets of school and university data separately (\Cref{tab:distributions}). The constant learning rate for the fine tuning is optimized on the dev sets, which yields $2 \times 10^{-4}$ for the school set and $5 \times 10^{-5}$ for the university set. 
Additionally, the same language models (LM) as in \cite{zhou2020rwth} are used for recognition, which are trained on the TLv2 LM training data. These include both the 4-gram and the LSTM-based LM, whose perplexity on both data sets is shown in \Cref{tab:perplexity}. To enhance the recognition output, we apply confusion network decoding by default.

One major challenge here is the large difference of voice characteristics between the experienced speakers in TLv2 and the young students in our data. 
This is especially the case for the school children whose vocal tract is still under development.
Speaker adaptation techniques are commonly applied to account for such speaker variability and to improve ASR performance. Following \cite{zhou2020rwth}, we adopt the vocal tract length normalization (VTLN) and the i-vectors-based speaker embedding approaches. As shown in Table \ref{tab:SA_WER}, VTLN gives the most improvement for the school set and i-vectors are more suitable for the university set. 

\begin{table}[h]
    \centering
        \caption{Perplexity of the TLv2 4-gram and LSTM LMs}
    \begin{tabular}{|c|c|c|c|c|c|c|}
        \hline
         \multirow{2}{*}{LM}& \multicolumn{3}{c|}{School}&\multicolumn{3}{c|}{University} \\ \cline{2-7}  
         &train&dev&test&train&dev&test\\
         \hline
        4-gram & 172 & 177 & 160 & 133 & 134 & 136\\
        \hline
        LSTM & 95 & 99 & 85 & 79 & 75 & 80\\
        \hline
    \end{tabular}
    \vspace{2mm}
    \label{tab:perplexity}
\end{table}

\begin{table}[h]
    \centering
        \caption{WER results of speaker adaptation (with a 4-gram LM)}
    \begin{tabular}{|l|c|c|}
    \hline
         \multirow{2}{*}{Model}& \multicolumn{2}{c|}{Dev} \\\cline{2-3}
         & School & University \\
         \hline
         baseline + fine tuning & 25.8 & 21.4 \\
         \hline
        \hspace{0.3cm}+VTLN & \textbf{23.3} & 18.4 \\
        \hline
        \hspace{0.3cm}+i-vectors & 24.0 & \textbf{18.2}\\
        \hline
        \hspace{0.3cm}+VTLN + i-vectors & 23.8 & 18.7\\
        \hline
    \end{tabular}
    \vspace{-3mm}
    \label{tab:SA_WER}
\end{table}

\section{Automatic Text Analysis (ATA) Setup}
Both manually and ASR generated transcripts of speech recordings were automatically analyzed using CoCoGen (short for \textit{Complexity Contour Generator}), a computational tool that implements a sliding window technique to calculate within-text distributions of scores for a given language measures (see e.g. \cite{kerz2020becoming,strobel2020relationship}, for recent applications of the tool in the area of language learning).The impetus for the implementation of the measures in the tool comes from a wealth of recent multidisciplinary research that adopts an integrated approach to language \cite{Christiansen2017} and language learning \cite{ellis2019essentials} as well as an extensive body of literature on CAF framework reviewed in Section 1. 
Here in this paper we employ a selection of 30 complexity measures (CMs) from a larger set of 91 CMs that fall into four categories (see below for the selection procedure): (1) syntactic CMs (N=12), (2) lexical CMs (N=11), (3) register-based n-gram frequency CMs (N=6), and (4) information-theoretic CMs (N=1).  CoCoGen uses the Stanford CoreNLP suite \cite{manning2014stanford} for performing tokenization, sentence splitting, part-of-speech tagging, lemmatization and syntactic parsing (Probabilistic Context Free Grammar Parser \cite{klein2003accurate}). The implementation of syntactic features follow the descriptions in \cite{lu2010automatic} and are based on the Tregex tree pattern matching tool \cite{levy2006tregex} with syntactic parse trees for extracting specific patterns. The operationalizations of measures of lexical complexity follow those described in \cite{lu2012relationship} and \cite{strobel2014:tracking}. The third group includes $10$ n-gram CMs that are derived from the five register sub-components of the COCA \cite{davies2008corpus}: spoken, magazine, fiction, news and academic language\footnote{The Contemporary Corpus of American English is the register-balanced corpus that contained 560 million words at the time the measures were derived.}. 
 The information-theoretic CMs, Kolmogorov complexity (KolDef), uses the Deflate algorithm \cite{deutsch1996deflate} to compress text and obtain complexity scores by relating the size of the compressed file to the size of the original file (see \cite{strobel2014:tracking} for the operationalization and implementation of these CMs). The CMs selection procedure was based on the following considerations: First, all CMs with near zero variance were removed. Second, redundant CMs were removed by considering the absolute values of all pair-wise correlations between the measures: If two variables had a correlation coefficient $r > 0.9$, we calculated the mean absolute correlation of each measure and removed the one with the largest mean absolute correlation.  

\section{Experimental Setup and Results}
To assess the impact of the transcription from a SOTA ASR system on automatic analysis of CMs in L2 spontaneous speech, Spearman correlation coefficients ($\rho$) were calculated between the results of all 30 speech measures based on manual transcription and the ASR output. In addition, we decided to zoom in on the school data set and examine which CMs discriminate most strongly among the three grade levels (10th, 11th, and 12th). This enables a more nuanced picture of the implications of using ASR for subsequent automated analysis of complexity. The reason for focusing on school data is to control for task type effects on linguistic complexity, since previous L2 research has demonstrated that topic and task type have a significant impact on measures of syntactic complexity and lexical sophistication (see e.g. \cite{alexopoulou2017task}). The relative importance of a given CM was quantified in terms of standardized coefficients of a mixed-effects ordinal regression model (cumulative link mixed model). Separate models were fitted to the scores from each of the 30 CMs to predict the cumulative probabilities for the outcome variable `school grade'.

\subsection{ASR Results}
The performance of the ASR system on the test set is provided in Table \ref{tab:ASR.results}. One speaker's recording from the university data had to be excluded for the analysis due to a strong distortion. The overall WER for the school data was 18.2\%, whereas the overall WER for the university data was 17.3\%. The most frequent recognition errors in the school data set were deletions (Del) (7.9\%), followed by substitutions (Sub) (7.3\%) and insertions (Ins) (3.0\%). In the university data, the most frequent recognition errors were substitutions (7.1\%), followed by deletions (6.7\%) and insertions (3.5\%). A closer look at the recognition error distributions revealed that the majority of errors involved mis-recognition of hesitation markers (such as \textit{uh}, \textit{uhm}) and monosyllabic grammatical words (such as \textit{a}, \textit{an}, \textit{and}, \textit{the}) (see \Cref{fig:FI} in the Appendix). As expected from the literature reviewed in Section 1, we observed substantial inter-individual variability among L2 speakers as shown by the speaker-specific ASR performance in Table \ref{tab:variability}. WER for different speakers can range from optimally 10\% to as high as 30\%.


\begin{figure*}[]
    \centering
    \includegraphics[width=1\textwidth]{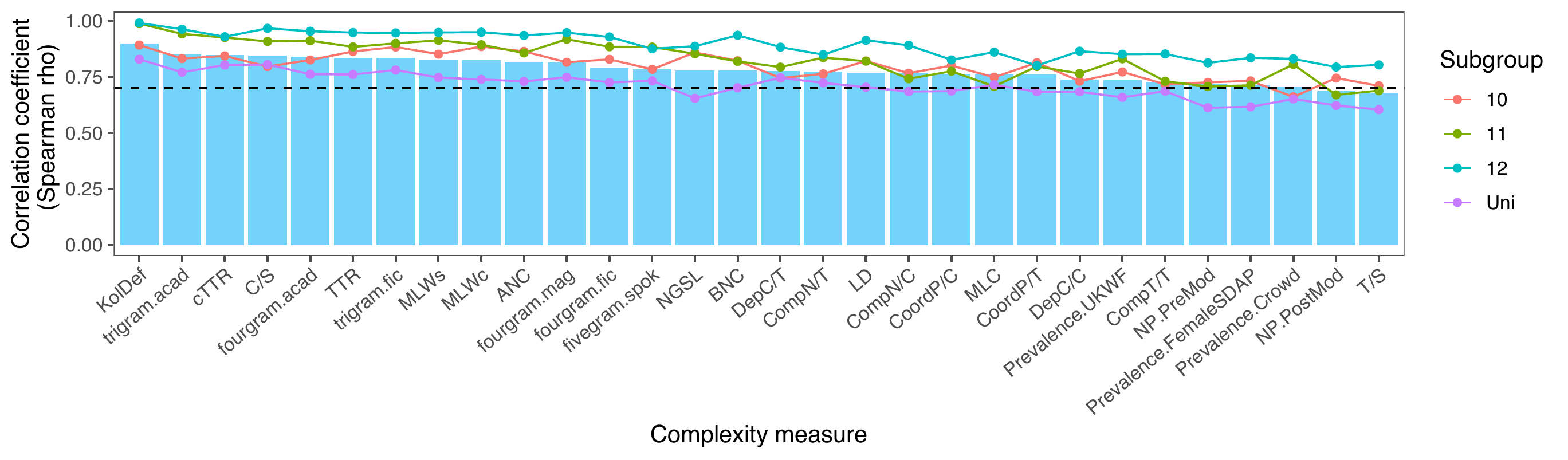}
    \caption{Correlation coefficients between CMs derived from manual vs. ASR-generated transcriptions. The height of the blue bars indicates overall correlation; colored lines represent the correlation coefficients by subgroup}
    \label{fig:Plotccorr}
    \vspace{-5mm}
\end{figure*}

\begin{table}[t!]
        \centering
          \caption{ASR performance on the school and university test sets (with an LSTM LM)}
        \begin{tabular}{l|cccc}
        \hline
             Test&Sub
             &Ins&Del &WER\\
             \hline
School&7.3&3.0&7.9&\textbf{18.2}\\
University&7.1&3.5&6.7&\textbf{17.3}\\
\hline
        \end{tabular}
        \vspace{1mm}
        \label{tab:ASR.results}
    \end{table}

\begin{table}[t!]
    \centering
    \caption{Statistics of L2 speaker-specific ASR performance (with an LSTM LM)} 
    \begin{tabular}{l|cccc}
    \hline
    &Mean&SD&Min&Max\\
    \hline
\textbf{School}&&&&\\
WER&18.4&5.2&8.4&34.7\\
Substitutions&7.3&3.0&2.7&18.1\\
Deletions&8.1&4.4&1.0&19.0\\
Insertions&3.0&1.4&1.0&7.2\\
\hline
\hline
\textbf{University}&&&&\\
WER&17.4&5.9&7.5&32.8\\
Substitutions&7.1&3.0&3.0&17.0\\
Deletions&6.9&3.3&2.8&15.5\\
Insertions&3.5&1.9&0.5&9.9\\
\hline
    \end{tabular}
    \vspace{-3.5mm}
    \label{tab:variability}
\end{table}

\subsection{Impact on ATA scores}
Complete descriptive statistics and results of the correlation analyses between CMs scores from manual vs. ASR-generated transcripts are provided in Table \ref{tab:corr} in the appendix
. Figure \ref{fig:Plotccorr} visualizes these correlations both overall and across data subgroups. Overall, the correlations between scores were high (average $\rho$ = 0.78, SD = 0.05). Strong correlations ($\rho > 0.7$) were found for 28 out of 30 measures. The remaining two measures exhibited moderate correlations with $\rho > 0.6$. The group of strongly correlated measures includes all register-based three-, four-, and five-gram CMs, all lexical complexity measures and the information-theoretic measure. For the syntactic CMs, we obtained the following results: All length-based CMs and all but one CM related to clausal subordination and coordination, such as number of clauses per sentence (C/S), mean clause length (MLC), or number of dependent clauses per T-unit (DepC/C), showed strong correlations. One phrasal CM (NP.PostMod) and one subordination CM (T/S) showed moderate correlations. 

Finally, we turn to the results of the ordinal regression models aimed at identifying those CMs that are most discriminative for the three school grades (see Figure \ref{fig:FI} in the Appendix). This was quantified on the basis of the sum of absolute threshold coefficients in these models.  The results indicated that the top eight most discriminative CMs include two academic (tri- and four-)gram measures, cTTR, two CMs pertaining to lexical sophistication (as gauged by word frequency lists from the ANC \cite{ide2001american} and, MLWs, the word length measured in syllables), two syntactic CMs (DepC/T and MLC) and one word prevalence measure (Prevalence.USAWF, for more details on these measures, see \cite{johns2020estimating}). These findings are consistent with those reported in the literature on automated approaches to L1 and L2 writing, which show a shift toward more advanced and sophisticated use of lexical items, including the increased use of academic vocabulary and multi-word sequences, across grade levels (\cite{durrant2019development}, \cite{kerz-etal-2020-becoming}). As reported above, the use of ASR to generate transcripts compared to manual transcripts did not have a large impact on the calculation of scores for seven of these eight CMs. However, at the same time, one CM (Prevalence.USAWF) of these top eight CMs showed only moderate correlations when calculated on ASR- vs. human-based transcripts.

\FloatBarrier
\section{Conclusion and Outlook}
Robust and reliable assessment of linguistic complexity in a second/foreign language is of particular importance for upper intermediate and advanced learner populations targeted in this paper. The importance of mastering complex and sophisticated L2 usage at higher proficiency levels is also reflected in the descriptors of the six CEFR levels introduced in Section 1. For example, in the section on qualitative aspects of spoken language use, the CEFR states that learners at B2 level should have ``a sufficient range of language to be able to give clear descriptions, express viewpoints and develop arguments without much conspicuous searching for words, using some complex sentence forms to do so". 

The first stage of the fully automated procedure for obtaining complexity measures (CM) of L2 speech performance is automatic speech recognition (ASR), which is required to convert the speech of L2 learners into a transcription that can serve as the basis for computing such measures. ASR of nonnative spontaneous speech is still challenging due to numerous factors that can affect its performance briefly outlined in Section 1 (for more details see \cite{zechner2020automated}). This raises the question of how the errors and uncertainties introduced by an ASR system affect the subsequent computation of text-based complexity measures. To address this question, we used a SOTA ASR system described in Section 3 in combination with automated text analysis (ATA) system (see Section 4). The latter was used to compute scores for a set of 30 selected CMs obtained on the basis of ASR-generated vs. manual transcriptions of 165 speech recordings of L2 speech produced by high school students (10th-12th grade) and 299 samples of spontaneous L2 speech produced by university students. The ASR system results with the overall WER of 18.2\% for the school data and 17.3\% for university data are very promising in light of the aforementioned challenges associated with the recognition of spontaneously produced L2 speech, including the significant inter-individual variability between L2 speakers that we also observed in both data sets. 
The degree of noise and uncertainty introduced by the ASR system had no adverse effect on the majority of CMs investigated in this paper, as indicated by high correlation coefficients. However, both the specific syntactic CMs pertaining to phrase-level complexity and CM of lexical sophistication pertaining to word prevalence were affected by ASR performance. 

In this paper we made a first important step towards systematically evaluating the impact of ASR on subsequent automatic analysis of linguistic complexity and sophistication in L2 speech. Although learners at more advanced levels can be expected to produce fewer grammar and vocabulary errors, we intend to consider the effects of such errors on automatic analysis of CMs in our future work. The importance of well-performing ASR is increasingly recognized in other contexts, such as automatic approaches to Alzheimer's disease (AD) detection \cite{zhou2016speech}.  A recent study \cite{balagopalan2019impact} investigated which speech patterns are most affected by certain types of ASR errors, such as word deletions and substitutions, and how this affects the performance of AD detection with machine learning models. In our future work, we will take a similar approach to investigate the different effects of error types on measures of syntactic and lexical complexity. 

\section{Acknowledgements}
\footnotesize
The authors acknowledge the financial support from the ERSSeed Fund project (OPEN Seed Fund Call 2019) CEASELESS– Chunk Learning and the Development of Speaking and Listening Fluency: Integrating Experimental and ComputationalApproaches funded by the Federal Ministry of Education and Research (BMBF) and the Ministry of Culture and Science of the German State of North Rhine-Westphalia (MKW) under the Excellence Strategy of the Federal Government and the Länder.

\let\normalsize\small\normalsize
\let\OLDthebibliography\thebibliography
\renewcommand\thebibliography[1]{
        \OLDthebibliography{#1}
        \setlength{\parskip}{-0.3pt}
        \setlength{\itemsep}{1pt plus 0.07ex}
}

\bibliographystyle{IEEEtran}
\bibliography{mybib}
\vfill\clearpage

\pagebreak
\section{Appendix}
\begin{table}[ht]
\centering
\setlength{\tabcolsep}{4pt}
\caption{Descriptive statistics for 30 CMs for manual vs. ASR-generated transcripts along with Spearman correlation coefficients ($\rho$)}
\label{tab:corr}
\begin{tabular}{|c|cc|cc||c|}
  \hline
      & \multicolumn{2}{c|}{\textbf{Manual}}& \multicolumn{2}{c||}{\textbf{ASR}}& \\ 

  CM & M & SD & M & SD & $\rho$ \\ 
  \hline
 KolDef & 0.93 & 0.54 & 0.91 & 0.36 & 0.90 \\ 
trigram.acad & 7.95 & 11.63 & 7.95 & 11.36 & 0.85 \\ 
cTTR & 3.83 & 1.08 & 3.83 & 1.09 & 0.85 \\ 
C/S & 2.52 & 3.17 & 2.46 & 3.15 & 0.85 \\ 
fourgram.acad & 0.84 & 1.90 & 0.83 & 1.84 & 0.84 \\ 
TTR & 0.85 & 0.14 & 0.86 & 0.13 & 0.84 \\ 
trigram.fic & 7.29 & 11.16 & 7.14 & 10.87 & 0.83 \\ 
 MLWs & 1.38 & 0.28 & 1.40 & 0.28 & 0.83 \\ 
 MLWc & 4.36 & 0.89 & 4.41 & 0.84 & 0.82 \\ 
 ANC & 0.18 & 0.14 & 0.18 & 0.15 & 0.82 \\ 
 fourgram.mag & 1.08 & 2.24 & 1.05 & 2.14 & 0.82 \\ 
 fourgram.fic & 0.88 & 2.17 & 0.83 & 2.07 & 0.79 \\ 
 fivegram.spok & 0.33 & 1.17 & 0.32 & 1.06 & 0.78 \\ 
 NGSL & 0.11 & 0.13 & 0.11 & 0.12 & 0.78 \\ 
 BNC & 0.36 & 0.16 & 0.36 & 0.17 & 0.78 \\ 
 DepC/T & 0.95 & 1.75 & 0.97 & 1.89 & 0.78 \\ 
 CompN/T & 1.97 & 2.37 & 2.05 & 2.70 & 0.77 \\ 
 LD & 0.48 & 0.15 & 0.48 & 0.15 & 0.77 \\ 
CompN/C & 0.99 & 0.96 & 0.99 & 0.98 & 0.77 \\ 
 CoordP/C & 0.25 & 0.48 & 0.25 & 0.48 & 0.76 \\ 
 MLC & 8.37 & 5.80 & 8.29 & 5.84 & 0.76 \\ 
 CoordP/T & 0.45 & 0.77 & 0.47 & 0.80 & 0.76 \\ 
 DepC/C & 0.34 & 0.35 & 0.32 & 0.35 & 0.74 \\ 
 Prevalence.UKWF & 12.59 & 1.46 & 12.74 & 1.12 & 0.74 \\ 
 CompT/T & 0.47 & 0.49 & 0.46 & 0.49 & 0.73 \\ 
 NP.PreMod & 0.78 & 0.91 & 0.77 & 0.81 & 0.72 \\ 
 Prevalence.FemaleSDAP & 4.05 & 0.43 & 4.10 & 0.29 & 0.71 \\ 
 Prevalence.Crowd & 2.17 & 0.26 & 2.19 & 0.21 & 0.71 \\ 
 NP.PostMod & 2.85 & 5.24 & 2.96 & 6.81 & 0.69 \\ 
 T/S & 1.06 & 0.67 & 1.04 & 0.58 & 0.68 \\ 
   \hline
\end{tabular}
\end{table}

\begin{figure*}[b!]
    \centering
    \includegraphics[width=1\textwidth]{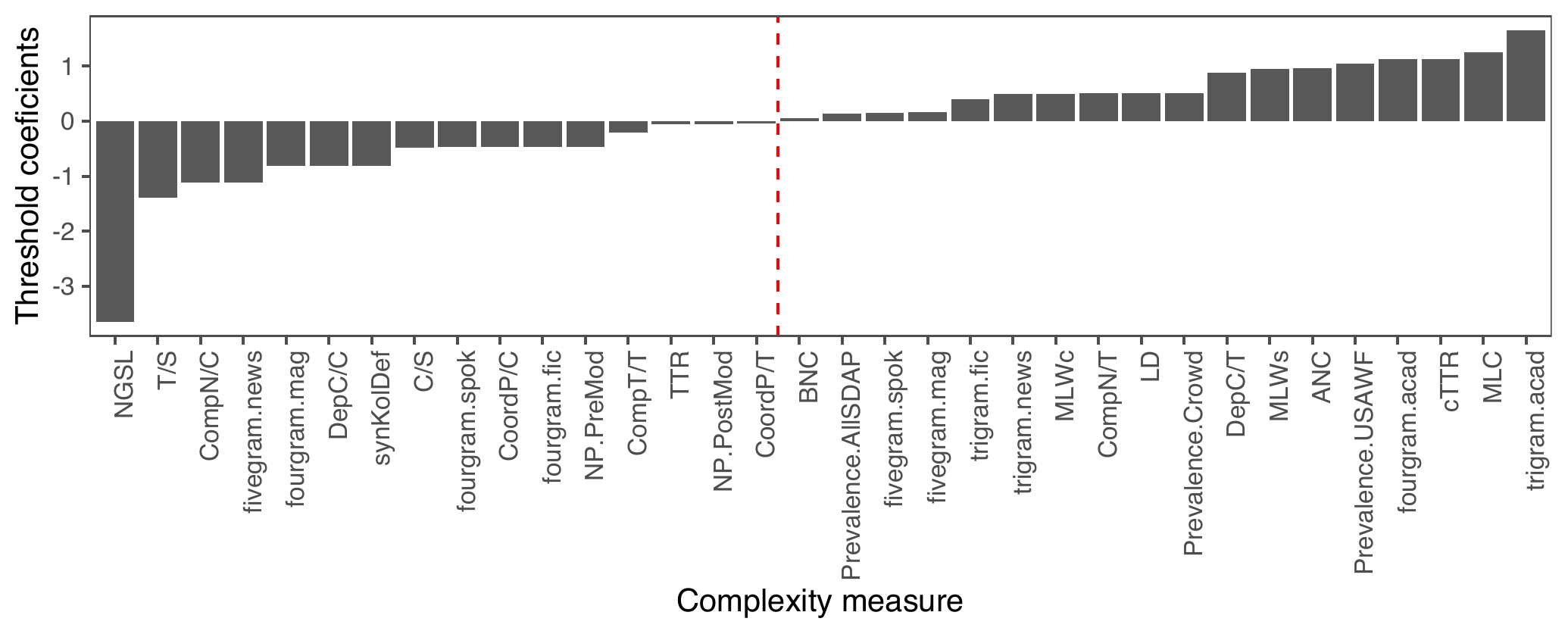}
    \caption{Feature importance (FI) as measured by the sum of absolute threshold coefficients (standardized). Features to the right of the red line have above-average FI scores.}
    \label{fig:FI}
\end{figure*}

\begin{figure*}
    \centering
    \includegraphics[width = \textwidth]{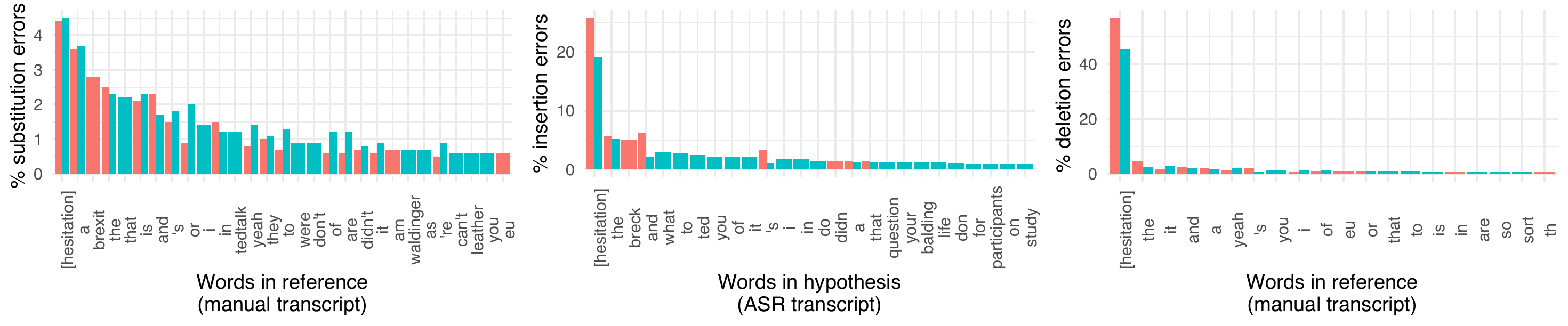}
    \caption{Percentage of substitution (left), insertion (middle) and deletion errors (right) across word types.}
    \label{fig:my_label}
    \vspace{-3mm}
\end{figure*}

\begin{table}[]
    \centering
    \setlength{\tabcolsep}{2pt}
    \caption{Mean linguistic complexity and sophistication scores (with standard deviations) across school grade levels 10-12}
    \begin{tabular}{|l|cc|cc|cc|}
    \hline
    &\multicolumn{2}{c|}{\textbf{Grade 10}}&\multicolumn{2}{c|}{\textbf{Grade 11}}&\multicolumn{2}{c|}{\textbf{Grade 12}}\\
    \hline
        &M&SD&M&SD&M&SD\\ 
        \hline
\textit{Syntactic complexity}&&&&&&\\
NP.PreMod&1.1&0.39&0.94&0.14&1.05&0.41\\ 
NP.PostMod&3.05&1.28&3.42&0.98&4.3&4.03\\ 
MLC&10.14&2.73&9.99&1.72&9.84&1.72\\ 
C/S&2.4&2.25&2.95&1.38&4.12&5.2\\ 
DepC/C&0.25&0.14&0.29&0.11&0.29&0.18\\ 
DepC/T&0.72&0.92&0.72&0.41&1.45&2.66\\ 
T/S&1.11&0.16&1.31&0.3&1.3&0.52\\ 
CompT/T&0.37&0.22&0.43&0.15&0.43&0.21\\ 
CoordP/T&0.73&0.53&0.64&0.35&0.66&0.46\\ 
CoordP/C&0.43&0.24&0.36&0.15&0.31&0.15\\ 
CompN/T&2.13&0.72&2.37&0.9&3.03&2.84\\ 
CompN/C&1.3&0.48&1.25&0.35&1.24&0.26\\ 
\hline
\textit{Information density (syntax)}&&&&&&\\
synKolDef&0.8&0.11&0.77&0.12&0.77&0.12\\ 
\hline
\textit{Lexical sophistication}&&&&&&\\
NGSL&0.15&0.06&0.15&0.05&0.15&0.04\\ 
ANC&0.2&0.06&0.22&0.04&0.2&0.05\\ 
BNC&0.39&0.06&0.4&0.05&0.38&0.05\\ 
Prevalence.Crowd&2.15&0.07&2.14&0.12&2.14&0.06\\
Prevalence.AllSDAP&4.66&0.16&4.67&0.13&4.7&0.07\\ 
Prevalence.USAWF&13.42&0.52&13.39&0.47&13.52&0.33\\
MLWs&1.44&0.13&1.45&0.07&1.42&0.09\\ 
MLWc&4.55&0.42&4.49&0.24&4.44&0.26\\ 
\textit{Lexical density}&&&&&&\\
LD&0.47&0.06&0.46&0.04&0.48&0.04\\ 
\textit{Lexical diversity}&&&&&&\\
TTR&0.83&0.06&0.8&0.06&0.81&0.07\\ 
cTTR&3.83&0.4&4.03&0.37&4.17&0.74\\ 

\hline
\textit{Register-specific n-gram use}&&&&&&\\
trigram.acad&10.58&5.1&10.16&4.97&14.75&14.87\\ 
fourgram.acad&1.34&0.79&1.15&0.9&1.37&0.9\\ 
trigram.fic&6.2&3.6&5.79&3.4&9.69&12.21\\ 
fourgram.fic&0.72&0.5&0.63&0.54&0.83&0.8\\ 
fourgram.mag&1.03&0.52&0.93&0.69&1.39&0.92\\ 
fivegram.mag&0.1&0.08&0.06&0.05&0.11&0.09\\ 
trigram.news&10.33&4.81&9.84&5.01&15.34&16.84\\ 
fivegram.news&0.11&0.08&0.09&0.08&0.12&0.08\\ 
fourgram.spok&1.62&0.81&1.64&1.21&2.35&2.12\\ 
fivegram.spok&0.22&0.2&0.24&0.22&0.31&0.32\\
\hline
    \end{tabular}
    \label{tab:grades}
\end{table}

\end{document}